\documentclass[10pt,twocolumn,letterpaper]{article}

\usepackage{iccv}
\usepackage{times}
\usepackage{epsfig}
\usepackage{graphicx}
\usepackage{amsmath}
\usepackage{amssymb}
\usepackage{caption}
\usepackage{subcaption}
\usepackage[labelfont=bf]{caption}
\usepackage{enumitem}
\usepackage{appendix}

\usepackage[breaklinks=true,bookmarks=false]{hyperref}

\iccvfinalcopy 

\def\iccvPaperID{29} 

\ificcvfinal\fi

\begin{document}

\title{MISFIT-V: Misaligned Image Synthesis and Fusion using Information from Thermal and Visual}

\author{
Aadhar Chauhan\\
University of Washington\\
{\tt\small aadharc@uw.edu}
\and
Isaac Remy\\
University of Washington\\
{\tt\small iremy@uw.edu}
\and
Danny Broyles\\
University of Washington\\
{\tt\small broyles@uw.edu}
\and
Karen Leung\\
University of Washington\\
{\tt\small kymleung@uw.edu}
}

\maketitle
\ificcvfinal\thispagestyle{empty}\fi

\begin{abstract}
   Detecting humans from airborne visual and thermal imagery is a fundamental challenge for Wilderness Search-and-Rescue (WiSAR) teams, who must perform this function accurately in the face of immense pressure. The ability to fuse these two sensor modalities can potentially reduce the cognitive load on human operators and/or improve the effectiveness of computer vision object detection models. However, the fusion task is particularly challenging in the context of WiSAR due to hardware limitations and extreme environmental factors. This work presents Misaligned Image Synthesis and Fusion using Information from Thermal and Visual (MISFIT-V), a novel two-pronged unsupervised deep learning approach that utilizes a Generative Adversarial Network (GAN) and a cross-attention mechanism to capture the most relevant features from each modality. Experimental results show MISFIT-V offers enhanced robustness against misalignment and poor lighting/thermal environmental conditions compared to existing visual-thermal image fusion methods. The code is available at GitHub.\footnote{\url{https://github.com/Aadharc/Visual_Thermal_Image_Fusion.git}}
\end{abstract}

\vspace{-\baselineskip}

\section{Introduction} 

Search and rescue teams worldwide have increasingly relied on uncrewed aerial vehicles (UAVs) equipped with visual and thermal imaging sensors to enhance the rescuer's ability to detect and locate lost or injured persons. Often, these operations take place in wilderness environments (i.e., wilderness search and rescue [WiSAR]) featuring hazardous and difficult-to-access terrain, and in various weather and lighting conditions which affect the quality of information obtained from visual and thermal modalities. Ultimately, these conditions affect the ability to detect the presence or absence of missing persons in the imagery. The detection task in a search and rescue mission, whether performed by human imagery analysts, UAV operators, or computer vision algorithms, is a fundamentally crucial function which can make the difference between mission success or failure.

Thus, developing algorithms that can detect humans more reliably and in adverse lighting and weather conditions is of utmost importance, and multimodal image fusion is an active area of research that offers many advantages for this application.

\begin{figure}[t]
   \centering
   \includegraphics[width=0.49\textwidth]{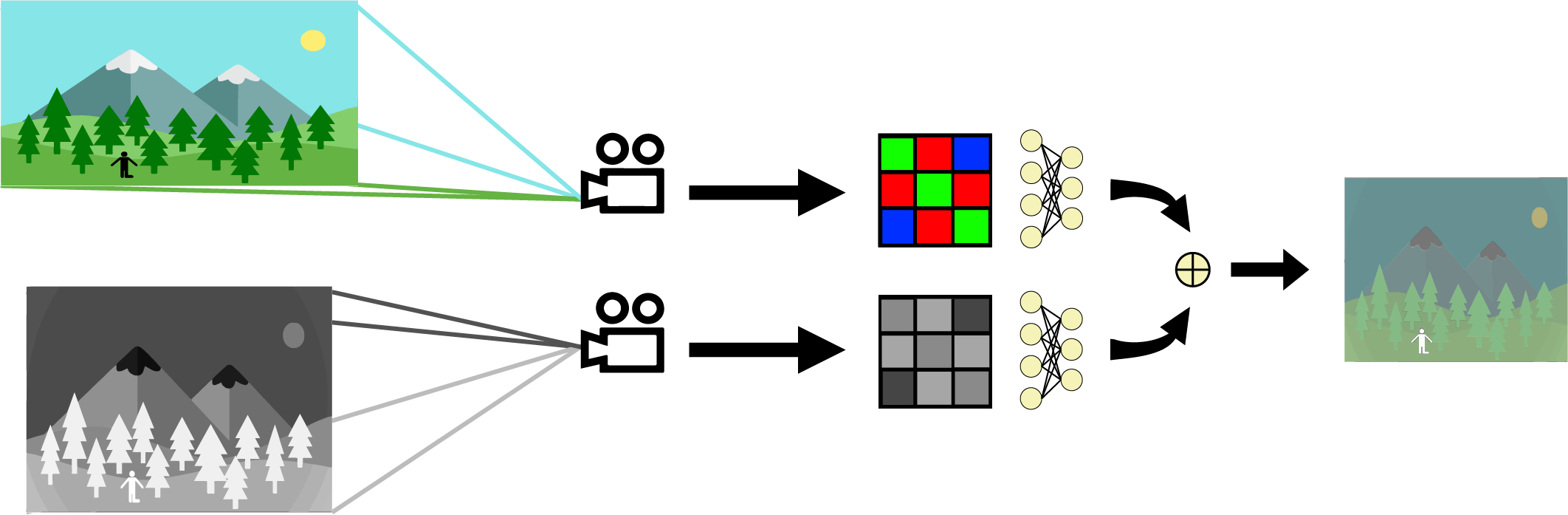}
      \caption{\textbf{System Conceptual Overview.} Images from visual (RGB) and thermal (IR) cameras capturing the same scene are misaligned due to physical properties.  MISFIT-V resolves this misalignment while emphasizing useful features from both modalities in the fused output.}
   \label{fig: problem form}
\end{figure}

Multi-modal image fusion is the act of combining information from multiple image modalities into a useful representation that can be more easily used for human or robotic perception-related tasks. This remains an important area of research with applications in medical diagnosis (fusing CT and MRI scans \cite{AzamKhanEtAl2022}), remote sensing (detecting environmental anomalies from satellite imagery \cite{XiuPanEtAl2022, YangWanEtAl2016}), and application featured in this paper, which is visual-thermal image fusion for enhanced human detection in UAV-aided search and rescue missions. However, achieving accurate and effective fusion between visual and thermal image modalities is particularly challenging in practice due to a variety of factors, including differences in sensor resolution, noise characteristics, and spatial and temporal misalignment which affect the image registration process. 

In this paper, we address the problem of fusing misaligned visual and thermal image pairs using unsupervised deep learning methods and provide qualitative and quantitative analysis of the resulting fused image quality.
The proposed method, Misaligned Image Synthesis and Fusion using Information from Thermal and Visual (MISFIT-V) accounts for practical considerations of WiSAR operations, and presents the following advantages:
\begin{itemize}[noitemsep]
    \item Image registration is not required 
    \item Ground truth fused images are not needed during model training 
    \item The fused output is human-interpretable 
    \item The fused output balances visual/thermal features
\end{itemize}

\subsection{Problem Statement}

The goal is to fuse a visual-thermal image pair $(I_\mathrm{RGB}, I_\mathrm{IR})\footnote{To ensure compatibility with the network architecture, the images undergo a preprocessing step that adjusts their sizes to be the same before feeding them into the network.} \in \mathbb{R}^{H \times W \times 3} \times \mathbb{R}^{H \times W \times 1}$, no necessarily aligned, into a single image $I_\mathrm{fus}\in \mathbb{R}^{H \times W \times 3}$ whereby features are aligned, and salient features from each image are combined. Here, $H$ \& $W$ represent the height and width of the corresponding images, and 3 \& 1 are the number of channels in respective images.
Motivated by the practical considerations of WiSAR operations, we further make the following assumptions in our problem setup:

\noindent\textbf{The image pairs are not aligned}: Given visual and thermal sensors function with different optical characteristics (e.g., field of view, resolution, and lens distortions), resulting in sets of misaligned visual and thermal images and largely uncorrelated feature sets. This makes traditional image registration/fusion techniques inadequate.

\noindent\textbf{Ground truth fused images do not exist}: Indeed obtaining fused images is the challenge, and manual image fusion is not a viable solution to obtain ground truth images.

\noindent\textbf{Fused image must be human-interpretable}: While the end goal is to use the fused image to aid in human detection, the fused image must remain human-interpretable given the human-on-the-loop nature of WiSAR missions, i.e., human operators monitor the video feed while a human detection algorithm runs concurrently.

\subsection{Organization} 

The rest of this paper is organized as follows. Section~\ref{sec:related work} discusses the relevant background and related work in image fusion. Section~\ref{sec:methodology} describes the architecture of MISFIT-V, and Sections~\ref{sec:results} and \ref{sec:conclusion} report the experimental results and conclusions, respectively.

\section{Related Work}
\label{sec: related work}

A vast number of approaches to the image fusion problem have been explored in the literature (see \cite{MaWangEtAl2023} for a review). We present a brief review on deep-learning-based fusion algorithms which have demonstrated state-of-the-art performance in this area.

\subsection{Visual \& Thermal Image Fusion}

Many standard image fusion methods for thermal and visual images, such as those proposed in \cite{AzarangManoochehriEtAl2019, FarahnakianHeikkonen2020, LiuChenEtAl2018, TangYuanEtAl2022}, rely on the input images that are aligned at the pixel level. This step of visual-thermal image alignment often requires precise hardware-level calibration and image processing, and misaligned features are the primary sources of error for these fusion algorithms. In many practical scenarios where visual and thermal cameras are used, the inherent differences between the two sensors (resolution, field of view, lens effects, and noise characteristics) make these images misaligned. 

Moreover, visual and thermal imaging sensors use distinctly different operating principles (visible spectrum light versus infrared radiation), often resulting in very little correspondence between each modality's features \cite{RicaurteEtAl2014}. 

\subsection{Multimodal Image Fusion with Generative Adversarial Networks (GANs)}

More recently, several methods (\cite{IsolaZhuEtAl2017, YangLiEtAl2020, ZhuParkEtAl2017}) propose using image-to-image translation as a way to address the lack of common features between visual and thermal images. Image-to-image translation is the act of taking images from one domain and transforming them so they have the style (or characteristics) of images from another domain. These methods \cite{WangLiuEtAl2022} use the Generative Adversarial Network (GAN) architecture as the backbone to translate the image from one modality to the other. However, the dataset \cite{ToetIJspeertEtAl1997} used for training the GAN architecture in these works is pre-registered and aligned at the pixel level, which can cause problems with respect to scalability. Since these methods require pixel-aligned multi-modal images, they cannot be utilized in many real-world applications where the alignment between visual and thermal sensor pairs is unknown a priori. For instance, a recent dataset, called WiSARD \cite{BroylesHaynerEtAl2022}, features visual and thermal image pairs taken from a UAV's perspective with annotations for human detection in a wilderness environment; however, the images are not perfectly aligned with each other on the pixel level.

\subsection{Cross-Attention Mechanism}

Natural language processing (NLP) research made substantial advancements in recognizing the relationships between various input sequence segments with the introduction of the Transformer model and the attention mechanism, and in particular the cross-attention mechanism is commonly used in deep learning models for tasks involving multiple modalities \cite{VaswaniShazeerEtAl2017}. Numerous studies have used a transformer model to successfully perform multi-modal data processing tasks, using a cross-attention module that enables the model to learn what information from one modality is relevant when analyzing features from another modality. For instance, \cite{ChenRichardEtAl2021} combined multi-scale picture characteristics using cross-attention, while \cite{BosePandeEtAl2021} combined visual and LiDAR image features using self-attention and cross-attention.

\section{Methodology}
\label{sec:methodology}

In this section, we described the proposed neural architecture of MISFIT-V, and motivate the loss function used to train the model.

\subsection{Proposed Architecture}

\begin{figure}[t]
   \centering
   \includegraphics[width=0.47\textwidth]{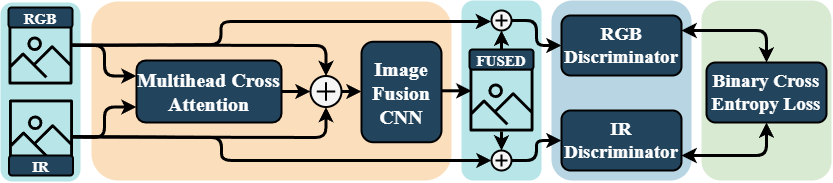}
      \caption{\textbf{MISFIT-V Training Pipeline.} Thermal (IR) and visual (RGB) images are fed into a generator (orange block) consisting of a cross-attention module and CNN to produce a fused image. The fused image is fed into both discriminator networks, encouraging a balanced set of features from both images.}
   \label{fig: proposed methods}
\end{figure}

Figure~\ref{fig: proposed methods} depicts our proposed model, which was inspired by the idea of using two discriminators presented in \cite{LiHuoEtAl2021, RaoXuEtAl2023}, leverages a Generative Adversarial Network (GAN) architecture as its backbone. The inclusion of two discriminators enables the preservation of information from both input modalities, ensuring that the distinctive features and characteristics of each modality are effectively retained. Next, we describe each component of the GAN architecture.

\subsubsection{Generator with Cross Attention}
\begin{figure}[t]
   \centering
   \includegraphics[width=0.47\textwidth]{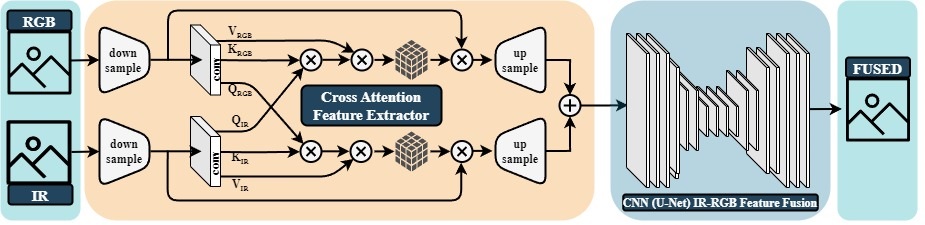}
      \caption{\textbf{MISFIT-V Generator Architecture.} Input images are fed into a downsampling CNN (`Down') separately to retrieve their features. These features are then fed into the cross-attention network to calculate the cross-attention map between the modalities, which are multiplied with the downsampled features and concatenated to form the input to the U-Net CNN, which generates a fused image.}
   \label{fig: generator}
\end{figure}

The generator, shown in Figure~\ref{fig: generator}, is designed to produce a fused image given a visual-thermal image pair. The generator network is comprised of two separate Convolutional Neural Networks (CNNs) for downsampling and feature extraction from the input images. Additionally, a Cross-Attention Network \cite{CaiZhuEtAl2022,VaswaniShazeerEtAl2017} is incorporated to capture meaningful and unique features from each modality, i.e., ``best of both worlds'', considering the diverse aspects that thermal and visible images focus on in the same scene (see Appendix~\ref{sec : cross attention}). The utilization of cross-attention eliminates the need for explicit alignment of the images, thereby addressing the fusion of misaligned input images. The outputs of the cross-attention network are a tuple of cross-attention maps with respect to both modalities, which are multiplied with the downsampled features and fed into an upsampling CNN. Finally, the outputs from upsampling CNNs are concatenated and fed into the U-Net which generates the fused image by integrating the information from both modalities to produce a more comprehensive and meaningful fused image (see Figure~\ref{fig: generator}).

\subsubsection{Dual Discriminators}

Given that no ground truth data exists, the proposed discriminator module is comprised of \textit{two} discriminator networks; one for visual and another for thermal, see Figure~\ref{fig: proposed methods}. 
Each discriminator takes a concatenated image comprised of the original (visual or thermal) and fused images and classifies them as either real or fake. In this way, the generator is discouraged from passing only the features from one modality through, so that it ultimately achieves a balanced set of features from each modality in the fused output.

\subsection{Loss Function}

Here, we describe the loss function used to train MISFIT-V. We employ an adversarial loss function to train the discriminators and generator in order to generate high-quality fused images.  The adversarial loss for the discriminator, which corresponds to a specific modality X (either thermal/IR or visual/RGB), is defined as follows:

\begin{equation}
\mathcal{L}_{\text{adv,X}} = -\log D_{\text{X}}(I_\text{X}) - \log(1 - D_{\text{X}}(I_{\text{fus}})),
\end{equation}
where $D_{\text{X}}(I_\text{X})$ represents the probability that $I_\text{X}$ is classified as modality X , and $D_{\text{X}}(I_{\text{fus}})$ represents the probability that $I_{\text{fus}}$ is classified as modality X by discriminator.


The generator loss is defined as the sum of the adversarial losses from both discriminators, weighted by hyperparameters $\lambda_{\text{IR}}$ and $\lambda_{\text{RGB}}$, respectively:

\begin{equation}
   \mathcal{L}_{\text{gen}} = \lambda_{\text{IR}} \cdot \mathcal{L}_{\text{adv,IR}} + \lambda_{\text{RGB}} \cdot \mathcal{L}_{\text{adv,RGB}}
\end{equation}

where $\lambda_{\text{IR}}$ and $\lambda_{\text{RGB}}$ control the relative importance of the respective losses in the overall generator loss. Here, the adversarial losses encourage the generator to create fused images that are indistinguishable from thermal images and visual images by training against two discriminators that try to classify them.

In addition to the adversarial losses, a Kullback-Leibler (KL) Divergence loss is used to compare the fused image generated by the generator with the original visual and thermal images in terms of their distribution. The KL Divergence loss is defined as:

\begin{equation}
   \mathcal{L}_{\text{KL}} = \text{KL}(I_{\text{fus}} || I_\text{IR}) + \text{KL}(I_{\text{fus}} || I_\text{RGB})
\end{equation}
where $\text{KL}(I_{\text{fus}} || I_\text{IR})$ and $\text{KL}(I_{\text{fus}} || I_\text{RGB})$ represent the KL Divergence between the fused image $I_{\text{fus}}$ and the thermal image $I_\text{IR}$, and between the fused image $I_{\text{fus}}$ and the visual image $I_\text{RGB}$, respectively.
Furthermore, an L1 loss is utilized to calculate the pixel-wise differences between the fused image and each of the original images. This loss can be expressed as:

\begin{equation}
   \mathcal{L}_{\text{L1}} = \|I_{\text{fus}} - I_{\text{IR}}\|_1 + \|I_{\text{fus}} - I_{\text{RGB}}\|_1,
\end{equation}
where $\|\cdot\|_1$ denotes the L1 norm.
The overall loss,

\begin{equation}
   \mathcal{L}_{\text{total}} = \mathcal{L}_{\text{gen}} + \lambda_{\text{KL}} \mathcal{L}_{\text{KL}} + \lambda_{\text{L1}} \mathcal{L}_{\text{L1}}
\end{equation}
is the sum of the generator loss, the KL divergence loss, and L1 loss weighted by hyperparameters $\lambda_{\text{KL}}$, and $\lambda_{\text{L1}}$ that control the relative importance of the KL divergence loss and L1 loss.


\section{Experimental Results}
\label{sec:results}

\subsection{Dataset and Training Details}

The model was trained using 2752 pairs of thermal and visual images from the WiSARD dataset \cite{BroylesHaynerEtAl2022}, with an 80:20 split for training and validation, and a separate test dataset of 200 pair of images was employed to evaluate the performance of the trained model. The network is trained for 20 epochs, using a learning rate of $1 \times 10^{-4}$. The hyperparameters for the training process are set as follows: $\lambda_{\text{KL}} = 10$, $\lambda_{\text{L1}} = 100$, $\lambda_{\text{IR}} = 1$, and $\lambda_{\text{RGB}} = 1$.

\subsection{Qualitative Analysis}

\begin{figure}[t]
   \centering
   \includegraphics[width=0.49\textwidth]{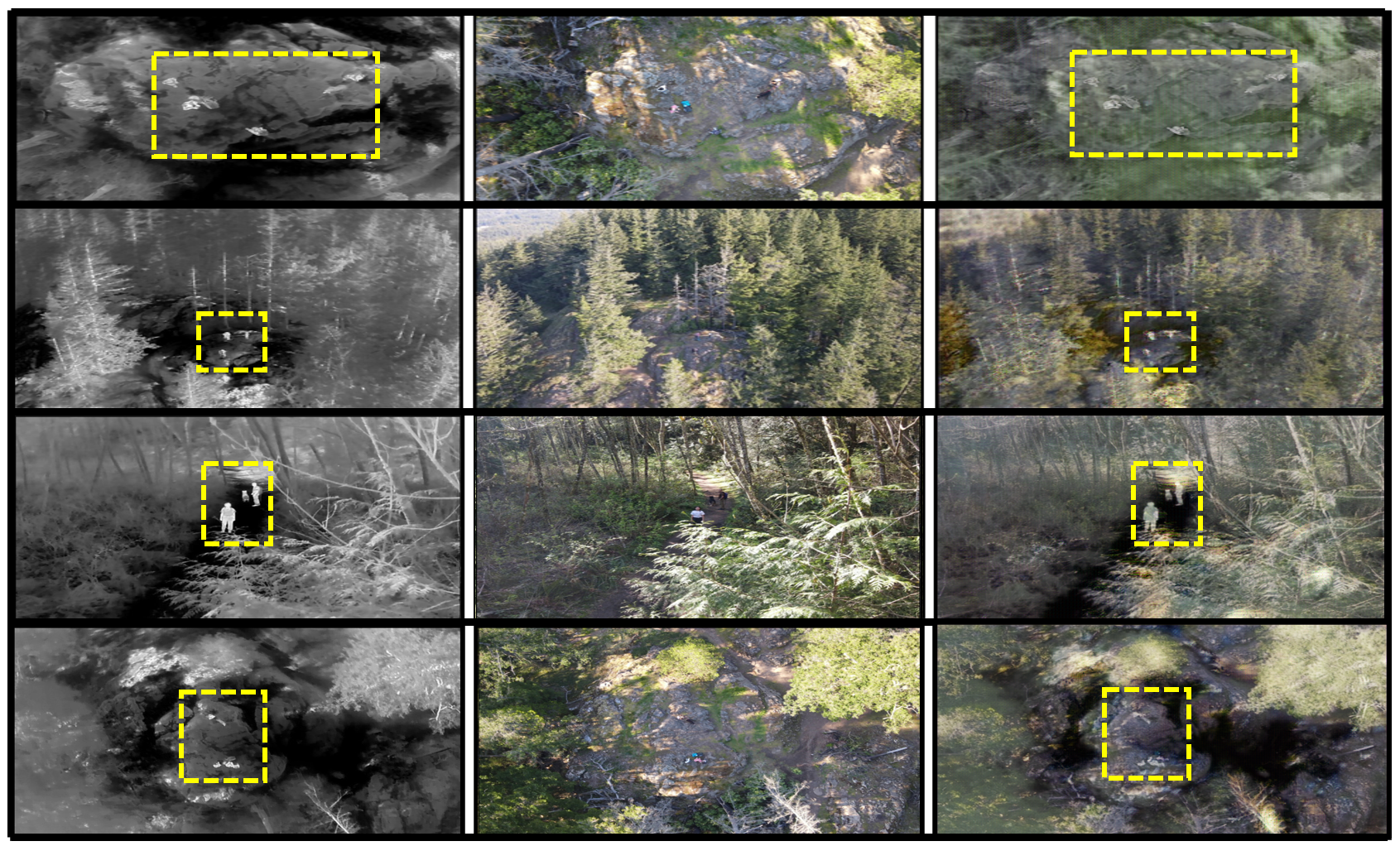}
      \caption{\textbf{MISFIT-V Results.} Each row, from left to right, shows a scene's thermal representation, its visual representation, and finally the resulting fused image via MISFIT-V, in a wilderness environment. The yellow bounding boxes highlight the locations of humans.}
   \label{fig: misfit_results}
\end{figure}

We demonstrate that the fused images provide a clearer representation of the environment than both modalities alone. From Figure~\ref{fig: misfit_results}, we see that our method produces well-fused images that retain the terrain features but extract the bright human silhouettes from the thermal image. However, our method still has limitations; for example, when objects of interest are small in size, the attention mechanism may encounter challenges in accurately determining the essential features from both modalities, leading to the emergence of ghost artifacts in the fused image (see the fourth row in Figure~\ref{fig: misfit_results}).

\subsection{Quantitative Comparison}

\begin{figure}[h]
   \centering
   \includegraphics[width=0.49\textwidth]{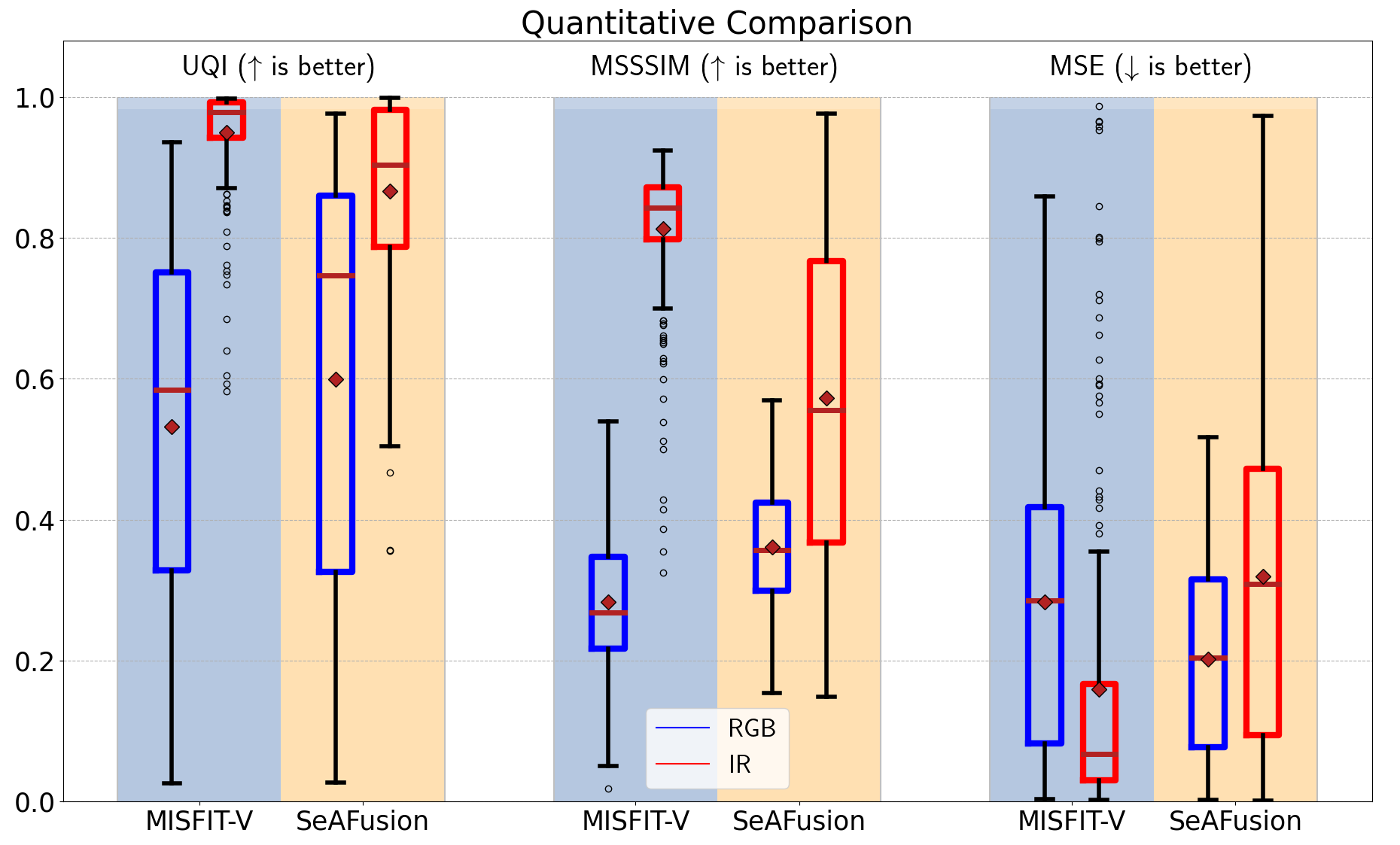}
   \caption{\textbf{Method Comparison.} This plot shows the results of three performance metrics comparing thermal and visual images against fused images generated by MISFIT-V and SeAFusion.}
   \label{fig: normal_plot}
\end{figure}

To quantify the fusion results, we evaluate the extent to which the information from each modality is preserved. This analysis involves calculating particular metrics individually for thermal and visual images against the fused image, to measure the level of information retention in the fusion process. We compare MISFIT-V against SeAFusion \cite{TangYuanEtAl2022}, another method that exhibits state-of-the-art performance for visual-thermal image fusion.
Given the formulation of SeAFusion, we had to evaluate both methods on an autonomous driving dataset \cite{HaWatanabeEtAl2017} which contains more structure than WiSAR settings and had ground truth labels. We compare against five metrics: Mean-squared-error (MSE), universal quality index (UQI), multi-scale structural similarity (MSSSIM), normalized mutual information (NMI), and the peak signal-to-noise ratio (PSNR). The y-axis of the comparison (see Figure~\ref{fig: normal_plot}) represents the numerical values corresponding to each metric. For brevity, we have presented the results for three metrics in this section which have been normalized. The plots for the remaining metrics can be found in Appendix~\ref{sec: Metrics}.
One interesting trend in our results is that MISFIT-V prioritizes information from the thermal modality to a greater extent and performs only marginally worse in retaining visual information when compared to SeAFusion. While MISFIT-V outperforms SeAFusion in some aspects and only slightly worse in others, it offers a significant advantage over SeAFusion by eliminating the need for semantic labeling and ground truth data, thus enhancing its scalability across diverse datasets. This characteristic enhances the applicability and adaptability of MISFIT-V in a wider range of scenarios.

\subsection{Ablation Study}

\begin{figure}[h]
   \centering
   \includegraphics[width=0.49\textwidth]{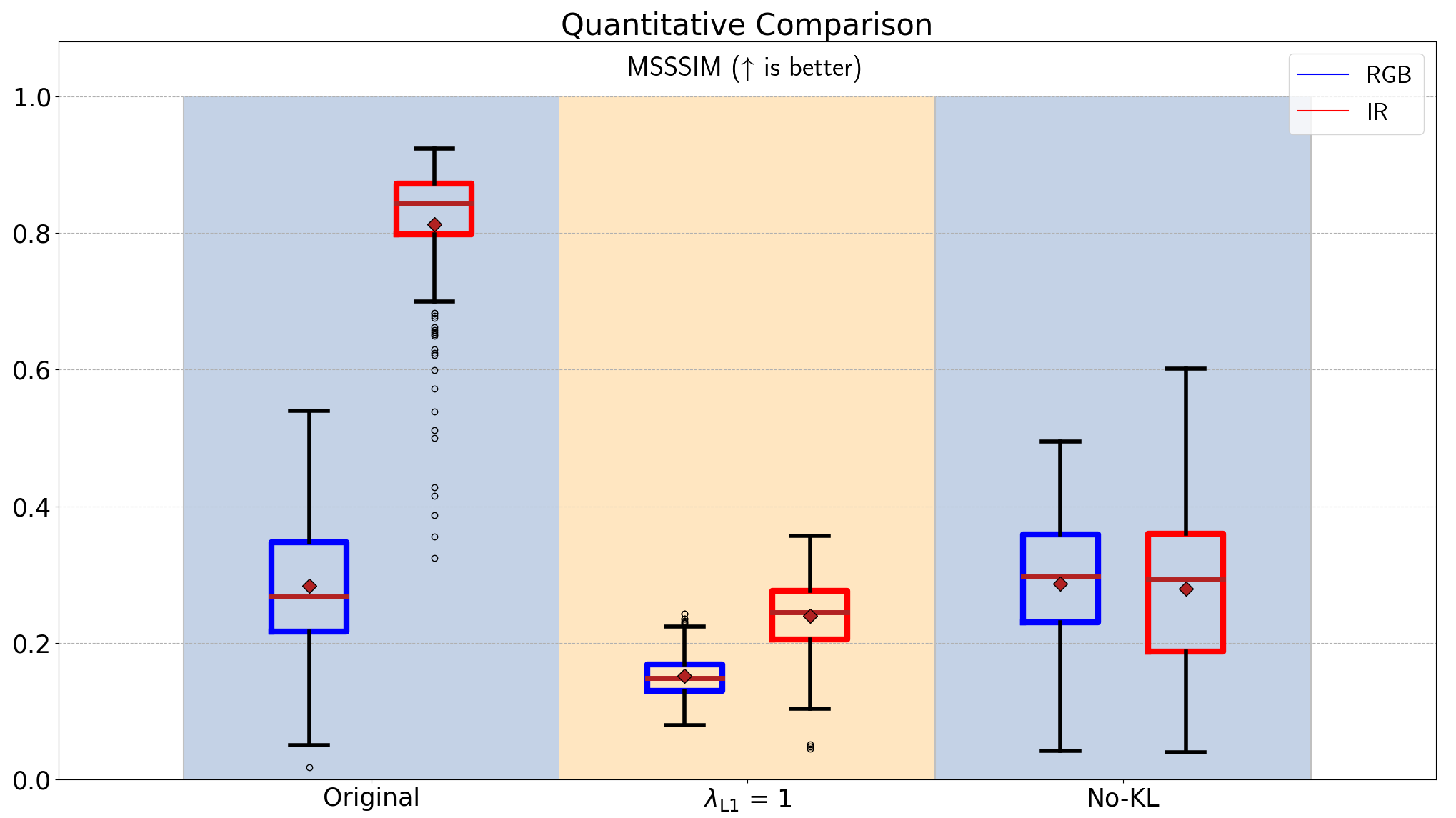}
   \caption{Comparison of image fusion performance using metrics `{MSSSIM}'. The first column, labeled as 
   `Original', presents the scores achieved by the original method. In the second column, labeled as 
   `$\lambda_\mathrm{L1} = 1$' the fusion performance is shown when the weightage of L1 loss is adjusted to 1. 
   The third column displays the results obtained when the KL loss term is omitted. Notably, the quality 
   of the fused image is observed to decrease when the KL loss is omitted, and this degradation is further 
   exacerbated when the $\lambda_\mathrm{L1}$ is set to 1. This visually emphasizes the significance of the 
   KL loss term and the weightage of L1 loss in maintaining the quality of the generated fused images.}
   \label{fig: Ablation_loss}
\end{figure}

In the pursuit of refining and optimizing our proposed methodology, we conducted 
an ablation study to meticulously analyze the effects of various modifications 
on the performance of our model. Through a series of controlled experiments, we 
aimed to dissect the contribution of specific components and choices within the 
architecture and total loss function. Here, we present the findings of our ablation study, comparing the 
original method with distinct variations: one involving the adjustment of the weightage of 
certain loss functions (Figure \ref{fig: Ablation_loss}) and excluding the cross attention mechanism between thermal and visual data.

\subsubsection{Impact of Loss Function Variations on Fused Image Quality}
\label{sec: ablation_study1}
An essential component of our proposed method revolves around the integration of L1 loss within 
the comprehensive loss function, aimed at optimizing the fusion procedure. To assess the significance 
of this specific loss term, we embarked on an experiment by modifying the weightage attributed to the 
L1 loss. In particular, we reduced the weightage from its original value of 100 to a much lower 
value of 1.  The rationale behind this manipulation was to ascertain whether diminishing the emphasis on 
the L1 loss would result in discernible alterations in the quality of the generated fused images. See Figure \ref{fig: Ablation_loss} and \ref{fig: Ablation_loss2}.

The outcomes of this experiment unveiled an intriguing insight. By reducing the L1 loss weightage, we 
observed a distinct deterioration in the comprehensibility of the resultant fused images. In other words, 
when the weightage was lowered from 100 to 1, the fused images exhibited a decrease in their interpretability 
and coherence. This phenomenon suggests that the L1 loss component indeed plays a pivotal role in shaping the 
clarity and visual coherence of the fused images. As such, its significance as a contributing factor to the 
overall loss function is highlighted, reinforcing the critical importance of its weightage within the fusion process.

In another variation, we examined the consequences of omitting the Kullback-Leibler (KL) loss 
term while maintaining the weightage of L1 loss at the original value of 100. This omission aimed 
to explore the repercussions of excluding the KL loss term on the final quality of the fused 
images. The subsequent analysis, as evidenced by the diverse metric plots presented below, 
offers valuable insights into the outcomes of these variations and their implications on the 
fused image quality. See Figure \ref{fig: Ablation_loss}. For additional information and graphical 
representations for other metrics, please refer to Appendix \ref{sec : Impact Loss Functions Variation} in this paper.

The experimental results shed light on an intriguing phenomenon. When the KL loss was removed from 
the loss function, we observed a discernible reduction in the quality of the fused images. This 
reduction was evident across various metrics that assess the image quality, underscoring the importance 
of the KL loss in enhancing the fusion process. By omitting the KL loss, which serves as a vital 
bridge between the latent space and the generated image, the model's ability to capture and reproduce 
intricate visual features was compromised. Consequently, the fused images exhibited a lower level of 
fidelity and coherence.

\subsubsection{Impact of Attention Mechanism on Fusion Quality}
\label{sec: ablation_study2}
The cross attention mechanism serves as a pivotal bridge between thermal and visual data, 
enabling the model to capture distinct yet complementary information from both modalities. 
In this ablation variation, we removed the cross attention mechanism entirely from the 
architecture to evaluate its influence on the fusion process.

\begin{figure}[h]
   \centering
   \includegraphics[width=0.49\textwidth]{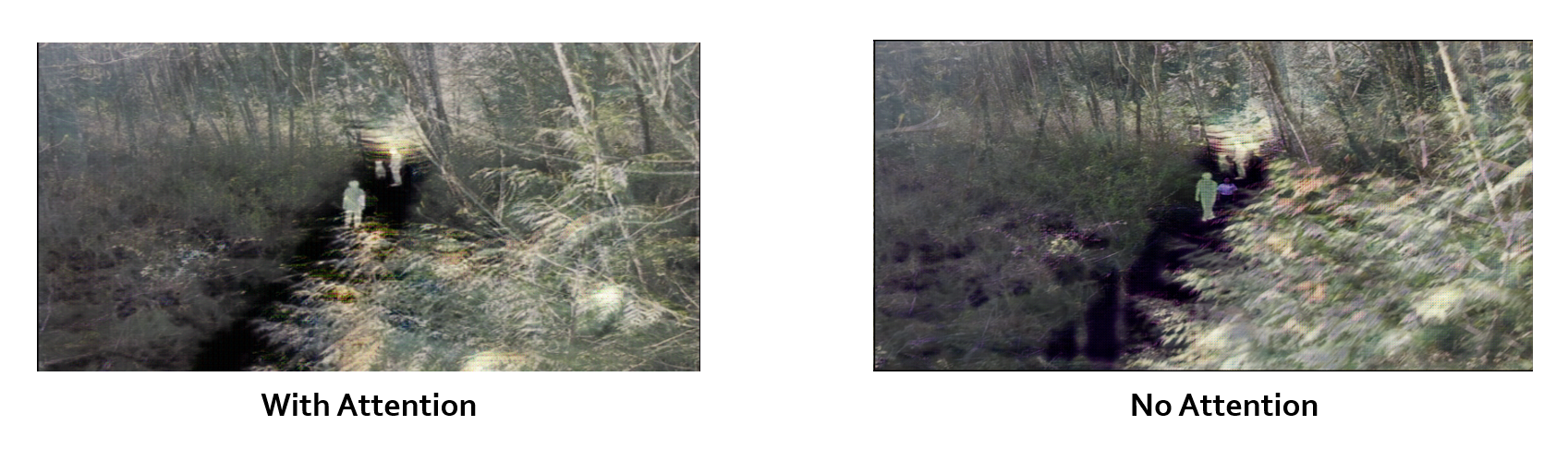}
   \caption{A comparison between fused images generated with and without the attention mechanism is 
   presented. The right side showcases the fused image generated without attention, where the inclusion 
   of visual features leads to the emergence of ghost artifacts. On the left side, the fused image 
   produced with the attention module is displayed, demonstrating that the attention mechanism effectively 
   omits less important visual features, resulting in a more coherent and comprehensible image.}
\label{fig: attn no attn}
\end{figure}

The comparison is presented in Figure \ref{fig: attn no attn}, where the right image displays the fused images generated without 
the attention mechanism and the left image showcases the fused images produced with the attention mechanism. 
Notably, the fused image generated without attention exhibited ghost artifacts and the inclusion of visual 
features, which can be attributed to a lack of emphasis on the essential characteristics of both modalities 
during the fusion process. In contrast, the fused images generated with the attention module demonstrated a 
distinct improvement in terms of quality and coherence. The attention mechanism effectively identifies and 
prioritizes significant features while minimizing the impact of less relevant visual features. This results 
in a more balanced and comprehensive fused image that accurately represents the salient information present 
in both thermal and visual modalities.

\begin{figure}[h]
   \centering
   \includegraphics[width=0.49\textwidth]{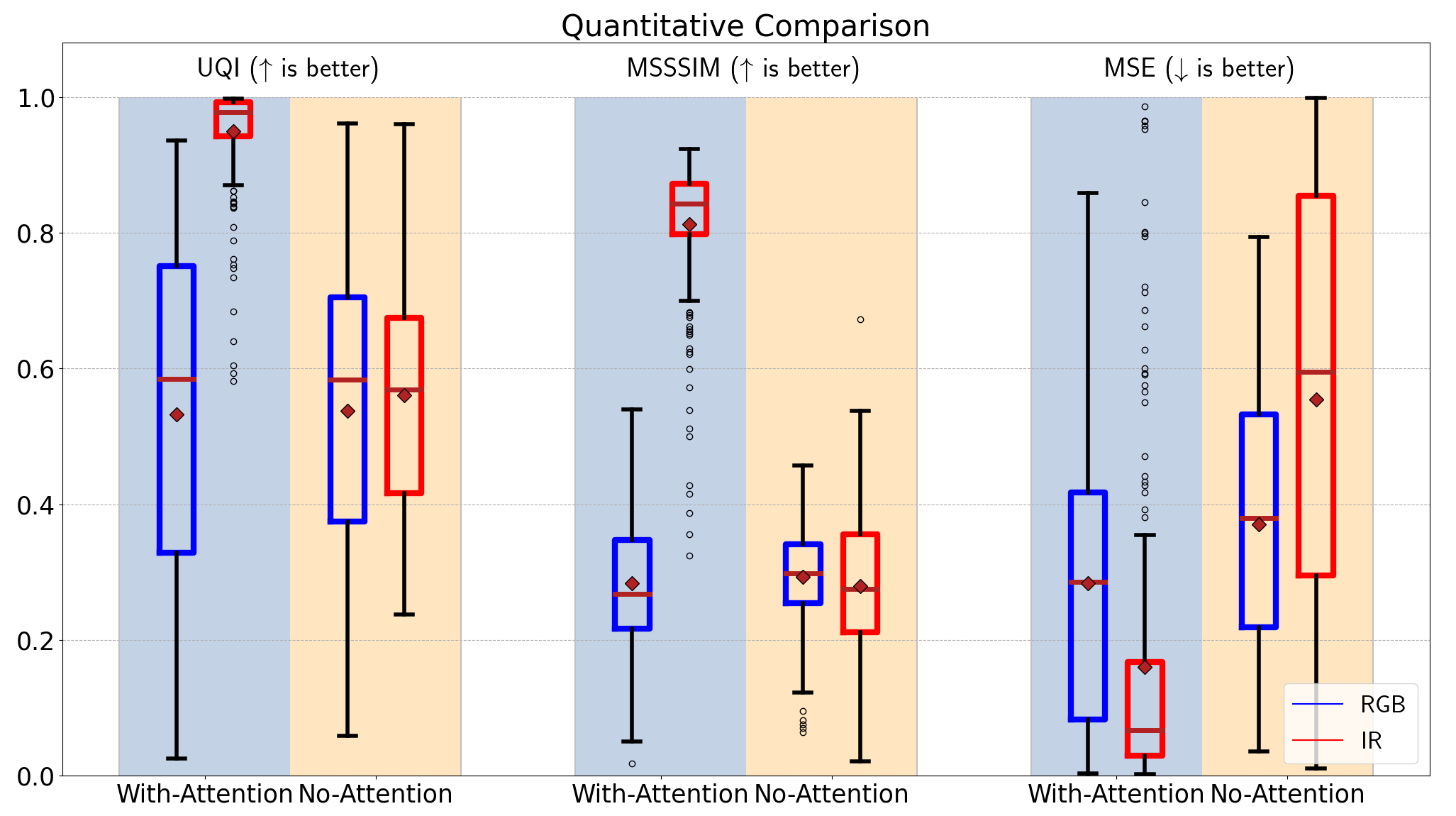}
   \caption{The plot illustrates the effect of utilizing the attention mechanism in the 
   image fusion process. In the columns named `With-Attention', fusion results obtained with the attention 
   mechanism are displayed, while in the columns named `No-Attention', results without the attention mechanism 
   are presented. Evidently, the quality of the fused images noticeably decreases when the attention 
   mechanism is not employed, underscoring its vital role in enhancing the fusion process by 
   selectively focusing on significant features from both modalities.}
\label{fig: ablation attention}
\end{figure}

The visual comparison is further supported by quantitative assessments, where metrics such as Mean-Squared Error ({MSE}), 
Multi-Spectral Structural Similarity Index ({MSSSIM}), Normalised Mutual Information ({NMI}), Universal Quality Index ({UQI}) and Peak Signal-to-Noise Ratio ({PSNR}) are employed to quantify the improvement in image 
quality achieved through the attention mechanism. Please refer to Figure \ref{fig: ablation attention} for a detailed comparison of metrics including {UQI}, {MSSSIM}, and {MSE}. For additional information and graphical 
representations for other metrics, please consult Appendix \ref{sec : Attention Mechanism Impact} in this paper.

\section{Conclusion}
\label{sec:conclusion}

We have presented MISFIT-V, a novel approach for visual-thermal image fusion leveraging a GAN architecture with two discriminators and a cross attention mechanism to blend crucial information from both modalities and hence eliminating the need to align the images altogether. The experimental results demonstrate the robustness and superior performance of MISFIT-V, outperforming a state-of-the-art baseline method while effectively handling misaligned images. By enabling a more complete representation of the environment, MISFIT-V has the potential to enhance WiSAR mission effectiveness and alleviate the cognitive load on human operators.

{\small
\bibliographystyle{ieee_fullname}
\bibliography{main}
}

\newpage
\appendix
\section{Cross-Attention}
\label{sec : cross attention}

\begin{figure}[h]
   \centering
   \includegraphics[width=0.49\textwidth]{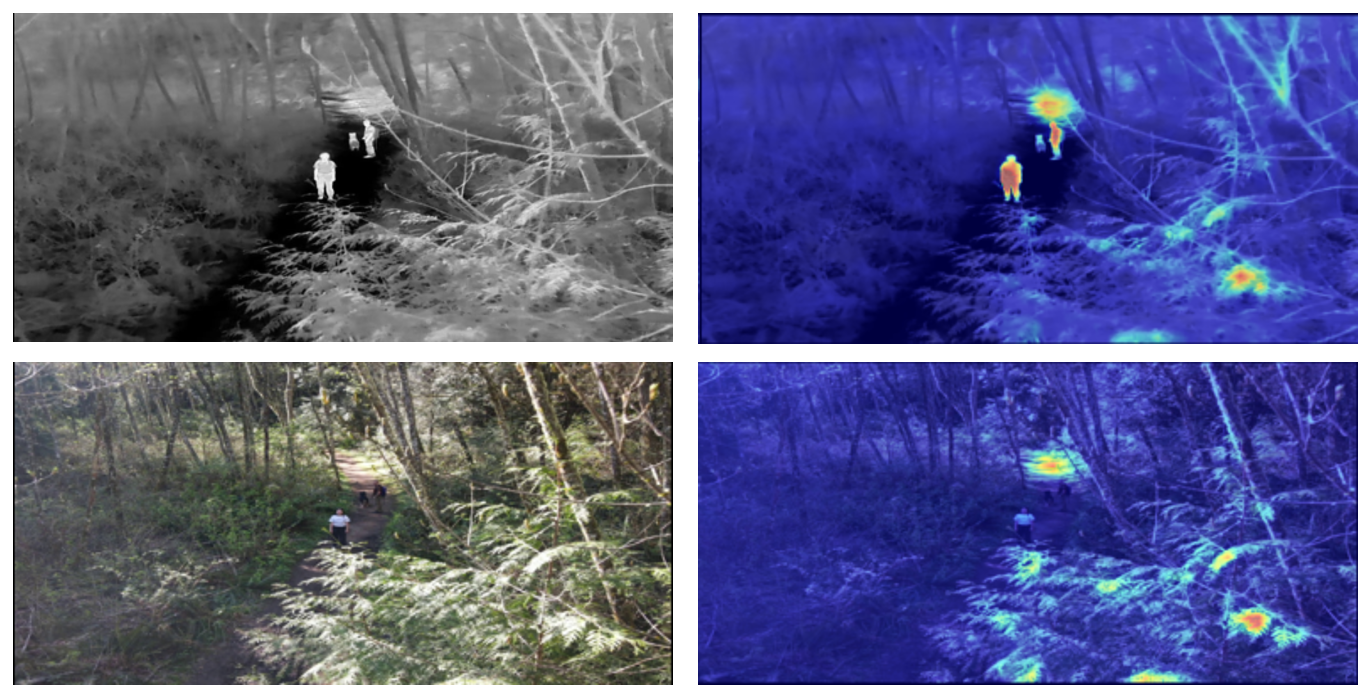}
      \caption{\textbf{Cross Attention}. The heatmaps on the right illustrate the regions of the images on the left that receive higher attention, with warmer colors indicating a greater degree of focus. In particular, areas with a reddish hue indicate heightened attention and prioritization.}
   \label{fig: attention_example}
\end{figure}

We incorporated cross-attention within our generator network to selectively focus on significant features from both image modalities. Cross-attention allows the generator to prioritize relevant information from each modality during the fusion process, enhancing the quality of the generated output. In the attention mechanism of the transformer architecture \cite{VaswaniShazeerEtAl2017}, queries, keys, and values are fundamental components. Queries represent the information that is sought or being searched for, while keys and values hold the information to be queried. In the context of cross-attention, queries from one modality are exchanged with queries from the other modality. We have visual and thermal modalities and queries from the visual modality are exchanged with queries from the thermal modality. (See Figure~\ref{fig: generator}).  This exchange enables the model to capture and leverage the complementary information from both modalities. By attending to important features in the thermal modality using visual queries, and vice versa, the generator can effectively combine and synthesize the relevant information from each modality for accurate fusion. See Figure~\ref{fig: attention_example}.

\section{NMI and PSNR Comparison}

\begin{figure}[h]
   \centering
   \includegraphics[width=0.49\textwidth]{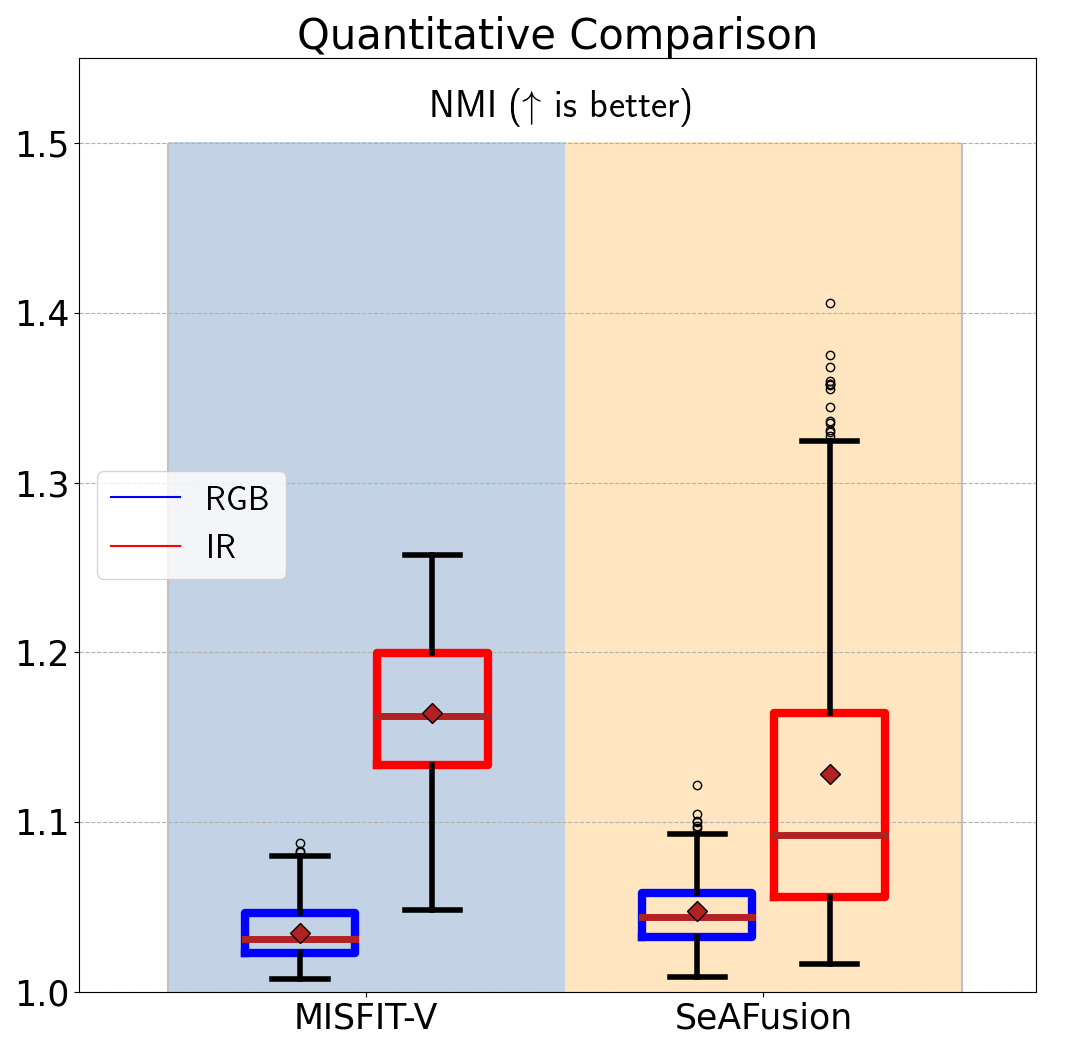}
      \caption{Normalized Mutual Information Comparison}
   \label{fig: nmi}
\end{figure}

\begin{figure}[h]
   \centering
   \includegraphics[width=0.49\textwidth]{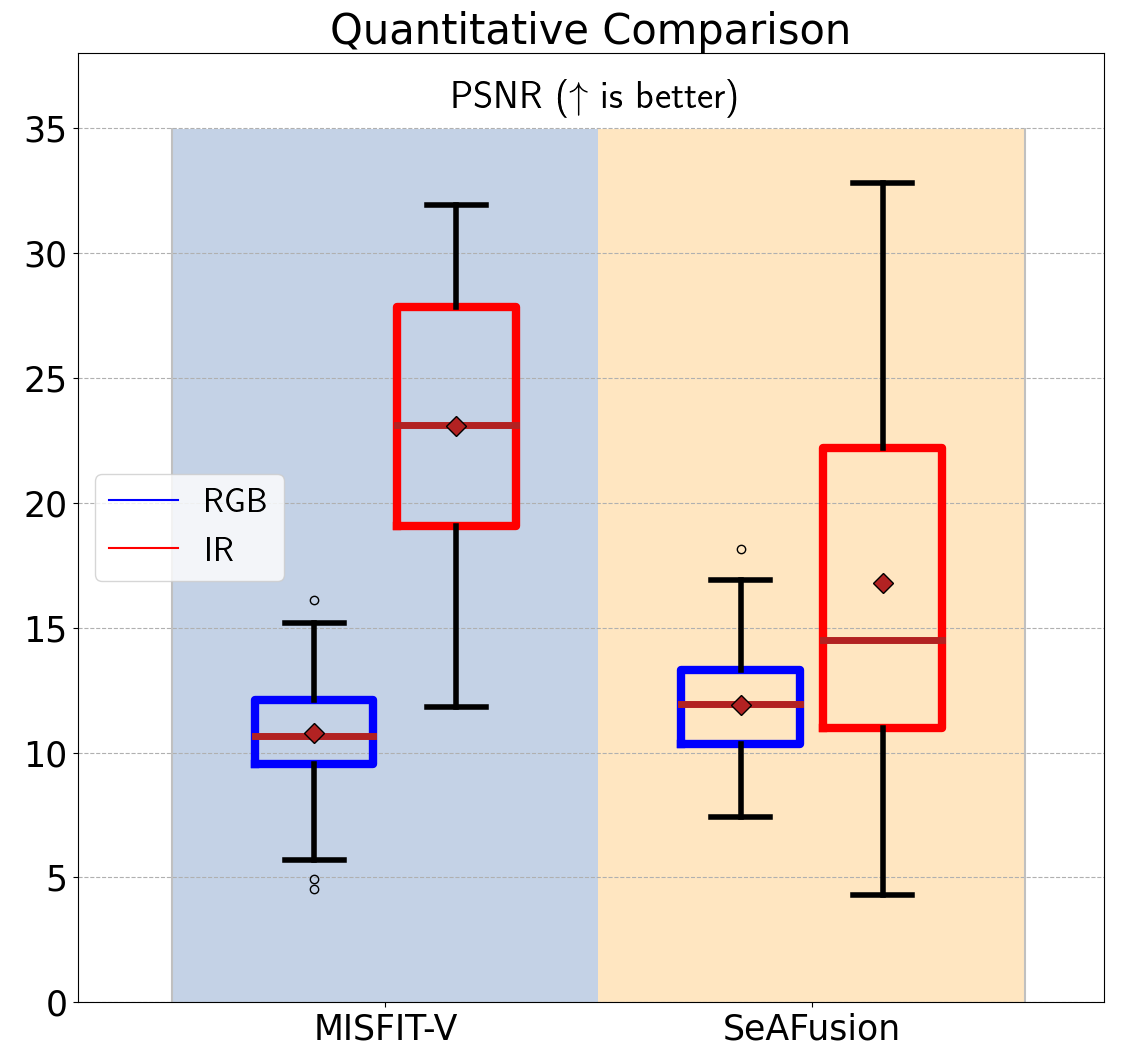}
      \caption{Peak Signal Noise Ratio Comparison}
   \label{fig: psnr}
\end{figure}

We present the results for image fusion using the metrics of Normalized Mutual Information (NMI) (see Figure~\ref{fig: nmi}) and Peak Signal-to-Noise Ratio (PSNR) (see Figure~\ref{fig: psnr}). The plots for NMI and PSNR provide a comprehensive evaluation of the image fusion performance for both thermal and visual modalities and they follow a similar trend as shown by the other three metrics in Figure~\ref{fig: normal_plot}.
\label{sec: Metrics}

\section{Impact of Loss Function Variations on Fused Image Quality}
\label{sec : Impact Loss Functions Variation}
\begin{figure}[h]
   \centering
   \includegraphics[width=0.50\textwidth]{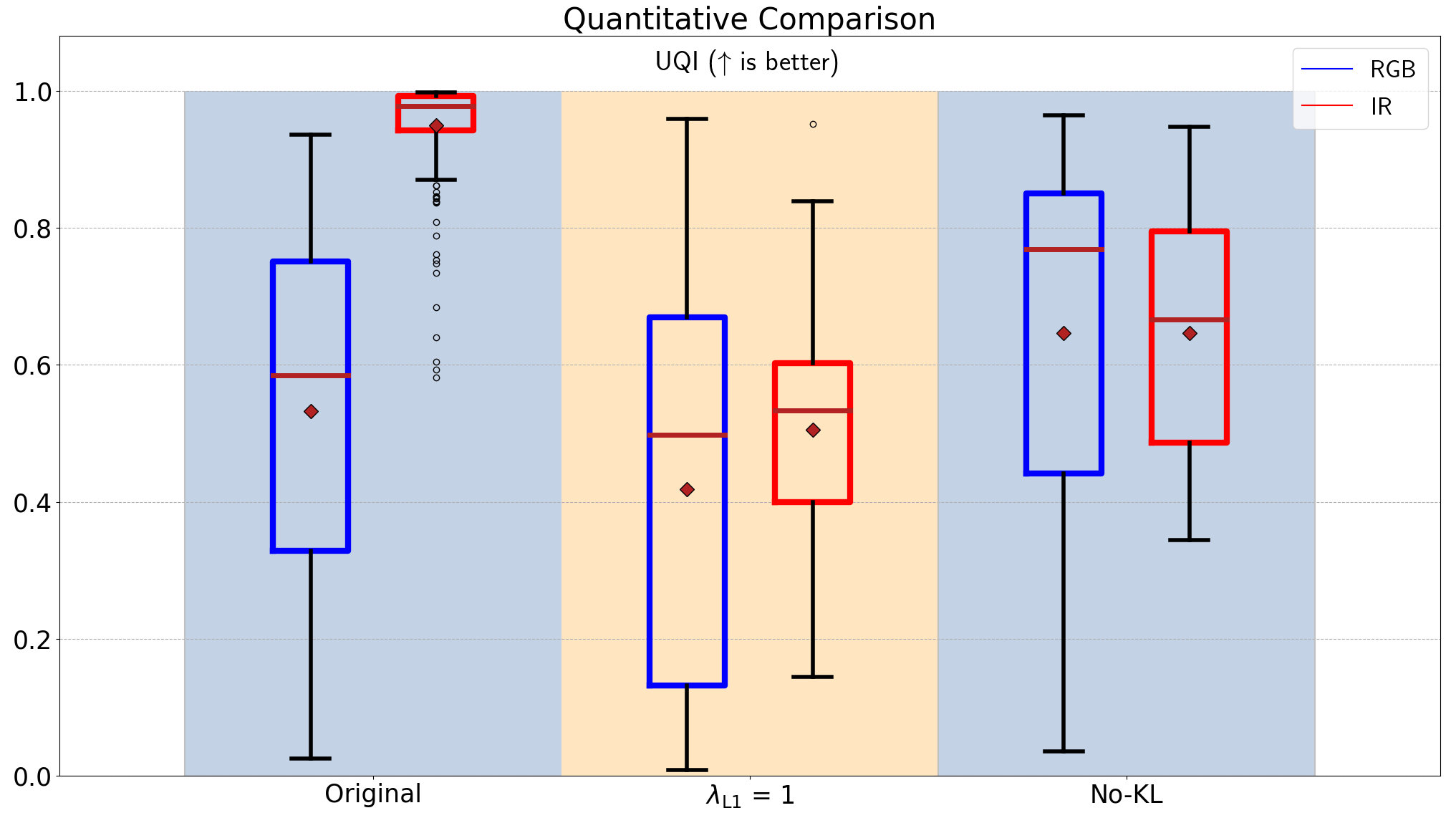}
   \caption{Comparison of image fusion performance using metrics `{UQI}'.}
   \label{fig: Ablation_loss2}
\end{figure}

\begin{figure}[h]
   \centering
   \includegraphics[width=0.50\textwidth]{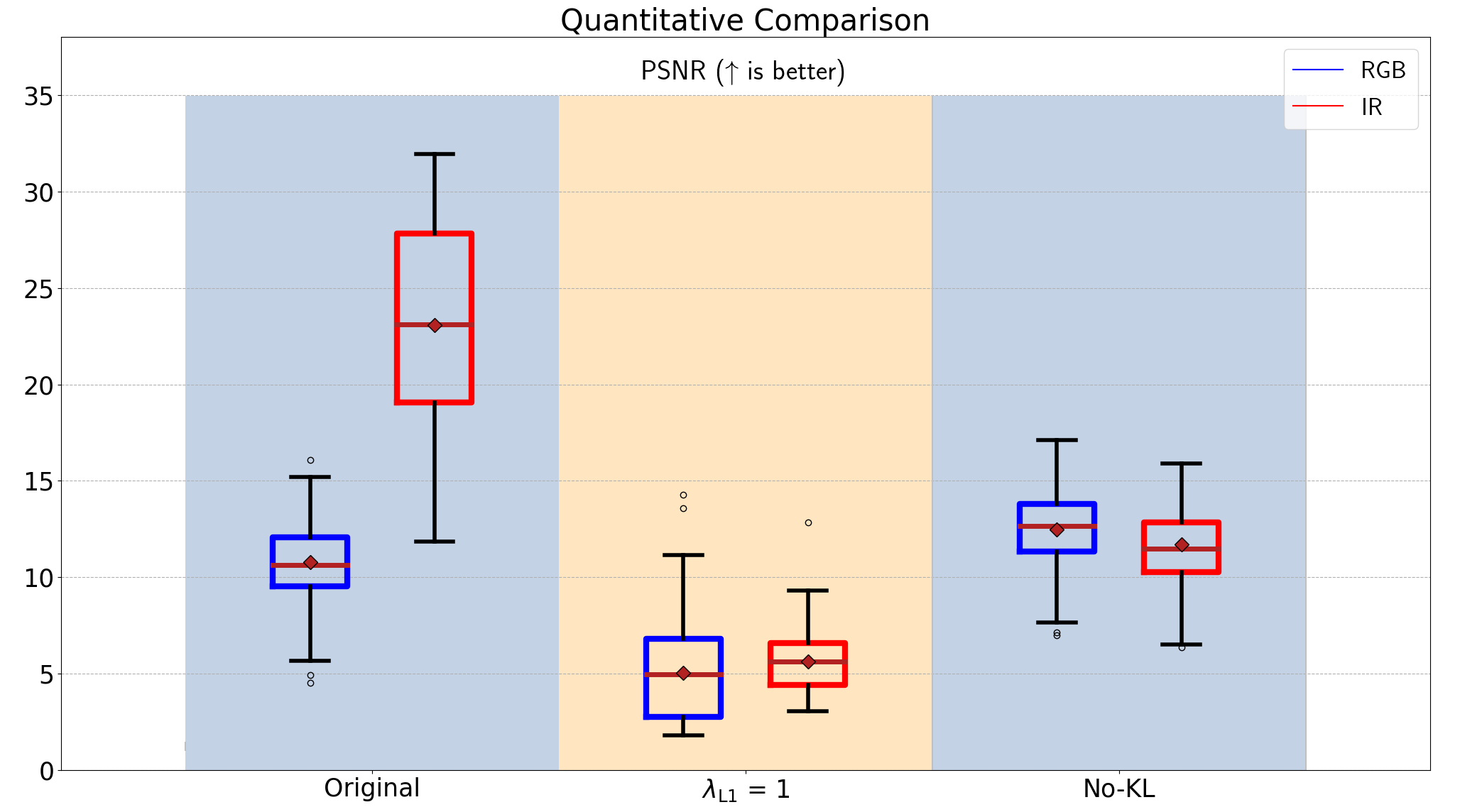}
   \caption{Comparison of image fusion performance using metrics `{PSNR}'.}
   \label{fig: Ablation_loss3}
\end{figure}

\begin{figure}[h]
   \centering
   \includegraphics[width=0.50\textwidth]{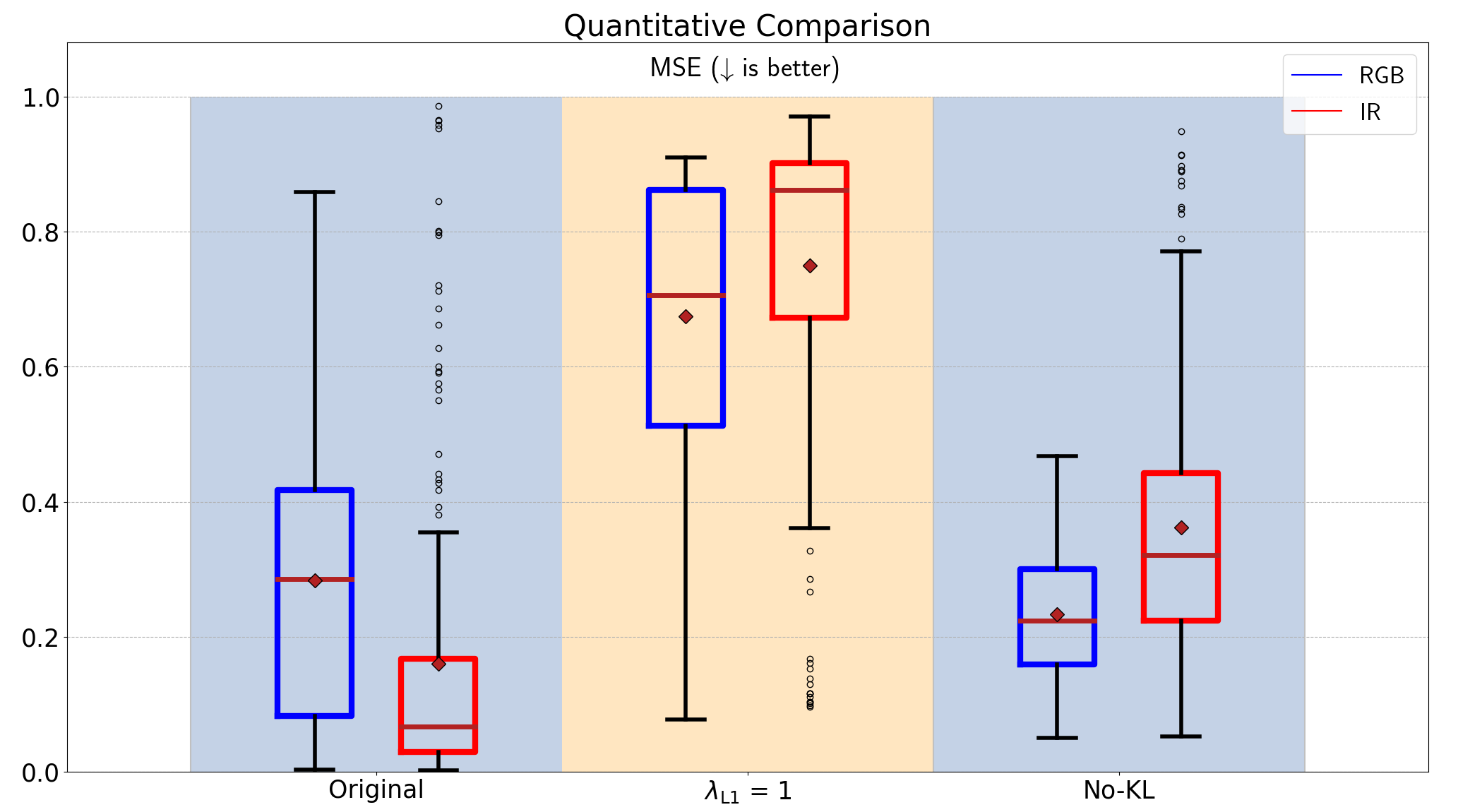}
   \caption{Comparison of image fusion performance using metrics `{MSE}'.}
   \label{fig: Ablation_loss4}
\end{figure}

In the appendix section, we delve deeper into our experimentation process, focusing on critical components of our proposed method. Specifically, we explore the significance of the L1 loss within our comprehensive loss function and the implications of omitting the Kullback-Leibler (KL) loss term as explained in \ref{sec: ablation_study1}. These insights provide valuable context for our research and shed light on the crucial factors influencing the quality of fused images. Please refer to Figure \ref{fig: Ablation_loss2}, \ref{fig: Ablation_loss3} and \ref{fig: Ablation_loss4} for visual representations of our findings in these experiments.

\section{Impact of Attention Mechanism on Fusion Quality}
\label{sec : Attention Mechanism Impact}
\begin{figure}[h]
   \centering
   \includegraphics[width=0.50\textwidth]{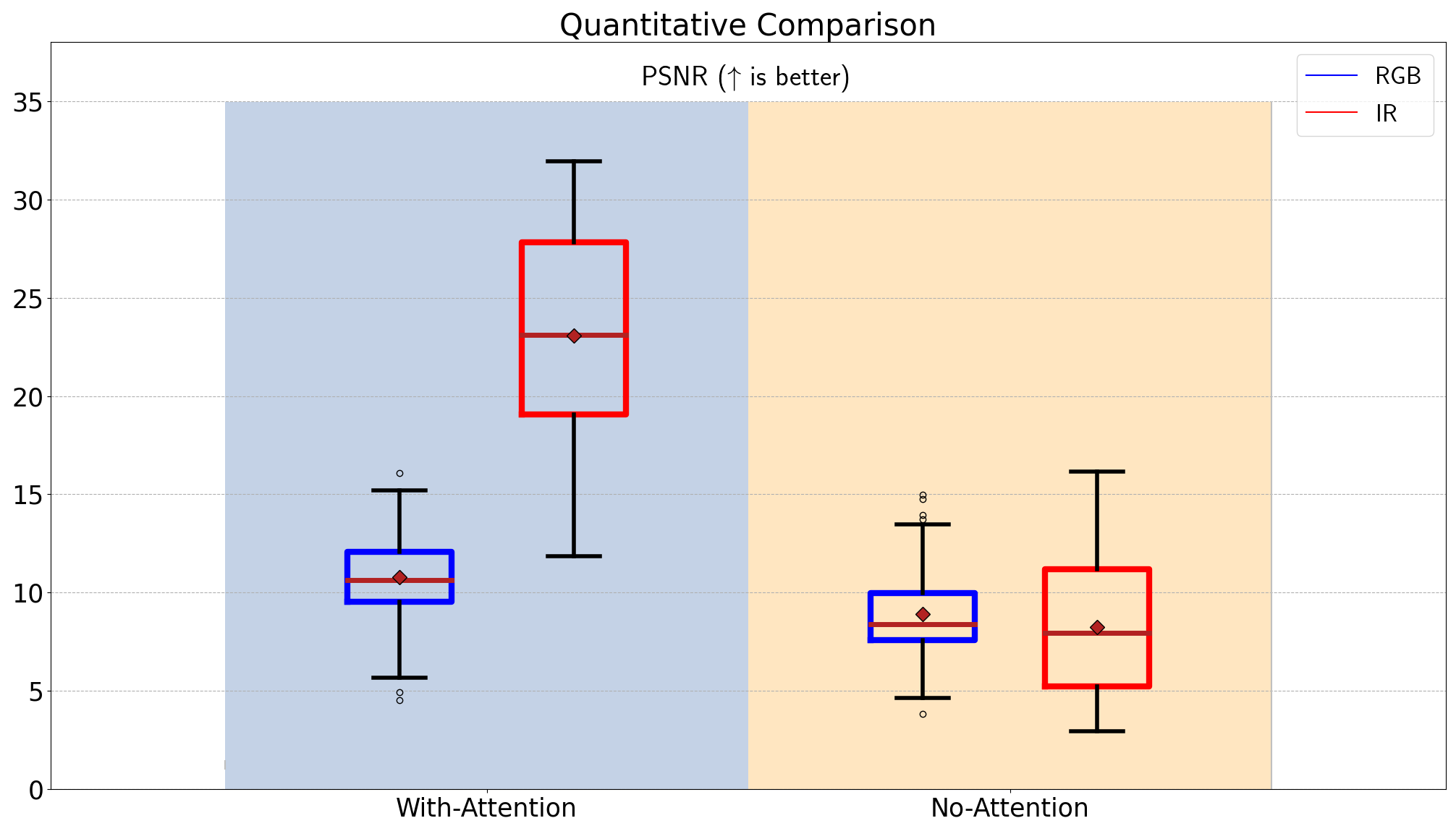}
   \caption{Comparison of image fusion performance using metrics `{PSNR}' to understand the impact of attention mechanism.}
   \label{fig: Ablation_loss5}
\end{figure}

\begin{figure}[h]
   \centering
   \includegraphics[width=0.50\textwidth]{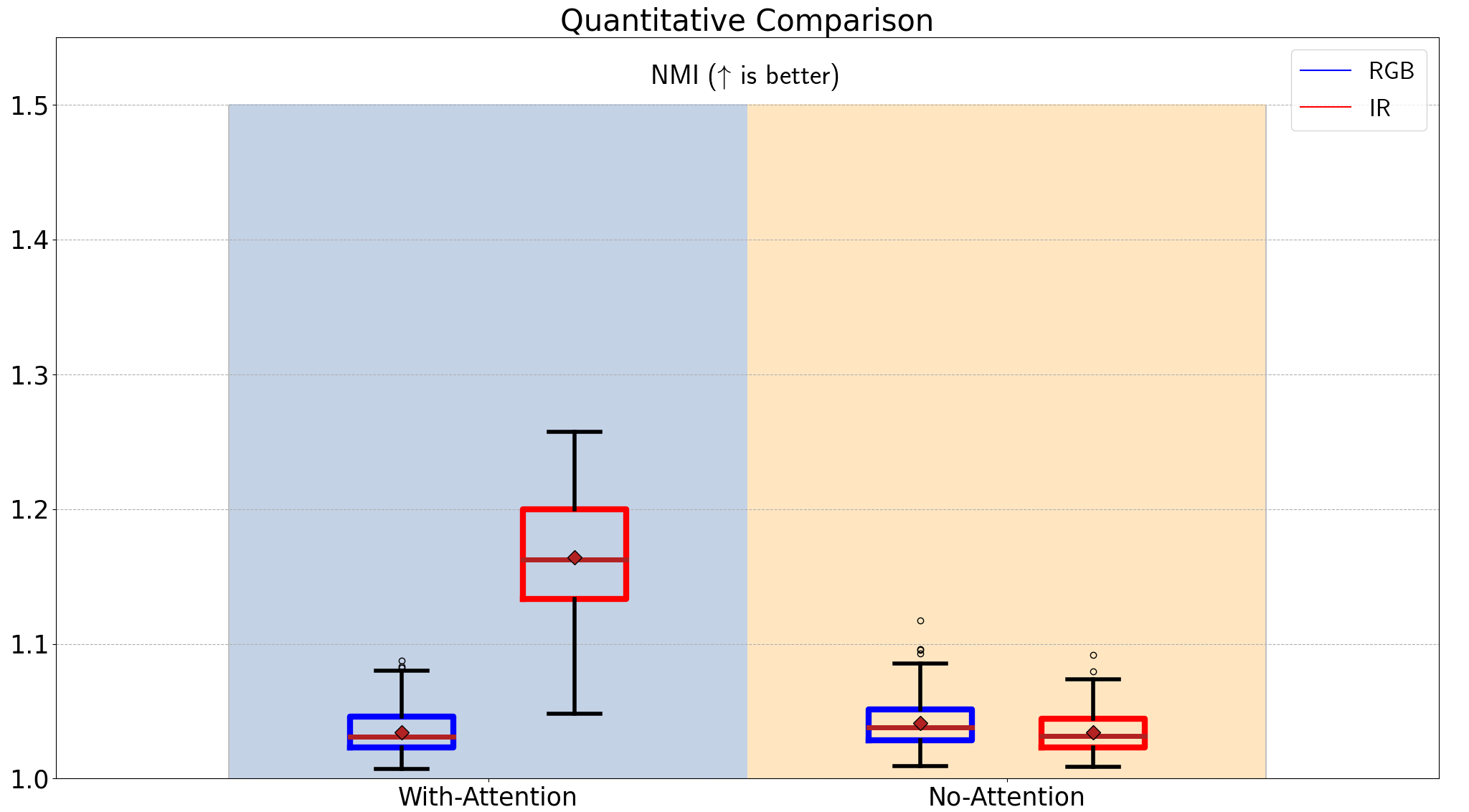}
   \caption{Comparison of image fusion performance using metrics `{NMI}' to understand the impact of attention mechanism.}
   \label{fig: Ablation_loss6}
\end{figure}

In the appendix section, we delve deeper into our experimentation process, focusing on critical components of our proposed method. Specifically, we explore the significance of the attention mechanism incorporated in the generator of our GAN, as explained in \ref{sec: ablation_study2}. These insights provide valuable context for our research and shed light on the crucial factors influencing the quality of fused images. Please refer to Figure \ref{fig: Ablation_loss5} and \ref{fig: Ablation_loss6} for visual representations of our findings in these experiments.

\end{document}